# Object Dimension Extraction for Environment Mapping with Low Cost Cameras Fused with Laser Ranging

Sachini Ekanayake[#1], Nilanga Thelasingha[*2], Bavantha Udugama[#3,] G.M.R.I. Godaliyadda[#4], M.P.B. Ekanayake[#5], B.G.L.T. Samaranayake[#6], J.V. Wijayakulasooriya[#7]

*Dept. of Electrical and Electronic Engineering, University of Peradeniya.*
*Peradeniya, Sri Lanka*

[1]emspe9353@gmail.com, [2] nilanga901@gmail.com,[3]bavantha11@gmail.com, [4]roshangodd@ee.pdn.ac.lk ,
[5]mpb.ekanayake@ee.pdn.ac.lk , [6]lilantha@ee.pdn.ac.lk , [7]jan@ee.pdn.ac.lk ,

*Abstract*-**It is essential to have a method to map an unknown terrain for various applications. For places where human access is not possible, a method should be proposed to identify the environment. Exploration, disaster relief, transportation and many other purposes would be convenient if a map of the environment is available. Replicating the human vision system using stereo cameras would be an optimum solution. In this work, we have used laser ranging based technique fused with stereo cameras to extract dimension of objects for mapping. The distortions were calibrated using mathematical model of the camera. By means of Semi Global Block Matching [1] disparity map was generated and reduces the noise using novel noise reduction method of disparity map by dilation. The Data from the Laser Range Finder (LRF) and noise reduced vision data has been used to identify the object parameters**

*Keywords*- **mapping, disparity, LIDAR, camera calibration, stereo vision, dimension extraction**

## I.  INTRODUCTION

Mapping is in general, graphical representation of environment with various objects and features. Mapping can be used to extract object information like height size, material types, object types etc. Machine vision is giving the capabilities of human vision to a machine using technology. human vision perceives as a colorful place with three dimensions.

Mapping plays a vital role since it allows us to respond to spacious geographic and social issues. Maps are useful for understanding and identifying spatial connections and explaining concepts in a visual manner that can be easily understood.

There are many applications in technology of environment mapping as military applications, disaster relief, navigation and exploration etc., this can be used in many security applications in industry as well. Simply places where human cannot access this will be an ideal solution.

 Explorations human inaccessible areas, finding dimensions of unknown terrains, by implementing this proposed method on a drone can be used for constructing 3D models of the down terrain. Further this can be extended to find material types and identifying object according to application. Number of sensors are available in the literature for detecting obstacles with help of SONAR [2], RADAR, LIDAR [3], vision systems and other proximity sensors [4].

 But many of them is running under high computational power and with high cost equipment.

Considering them stereo cameras are low cost and mapping has been done through stereo vision Path planning techniques for traversing given point with minimum travel cost has been developed for some time [5]. Also, Laser Range Finder(LRF) is an accurate device for measurement. Various applications have been done with LRF related to the title.

Hence through the proposed method we fused LRF data and stereo cameras with low computational cost to extract dimensions of objects while LRF is focused to find plane X, Y dimensions and stereo camera is used to find Z dimension. Altogether mapping of an environment can be achieved.

## II.  PROPOSED SOLUTION

 In the proposed solution, two-low cost off the shelf web cameras has been used for stereo vision and a simple neato $XV-11$ LRF salvaged from Neato vacuum cleaner as sensors. Since camera calibration is a requirement it was calibrated properly up to 14 distortion coefficients without being limited to usual method.

 From left and right images disparity map was obtained and the height of the object a found with it. Regarding the length and the width we use the LRF introducing this to find the planet of the objects and constructing the 2D environment as discussed in section [E]. method as  There are many [6] researches which have been done using stereo cameras but, taking multiple realization is a method with much computational cost in terms of practical implementation on a suitable platform. Hence, what has been proposed is from stereo cameras we extracted the object information of height and to construct the 3D object LRF is used to make the 2D environment. On top of that we are going to construct the 3D object. For this purpose, only the two left and right images are needed with object height information and the LRF sweep information. This is not much computationally costly or unrealizable.

 The cameras used has its own distortions. Radial distortions and tangential distortions are prominent two of those. Hence, it is necessary for them to be calibrated and undistorted first. Calibrating in the sense identifying the mathematical model and details of the camera, is done in two parts. First two cameras have been treated separately and calibrated and the error parameters were found. Also for the optimum disparity map generation rectification is necessary as the distortions are corrected and the images are aligned in a common horizontal line, so that the stereo matching can be executed.

Considering the height extraction, maximum intensity value of the disparity image was considered. Hence to remove the effect of ground plane noise values a constant distance was always maintained when taking images. So that identifying the maximum intensity with the relation of pixel – cm relation height of the object was found. There is a dimension change when taking 3D to 2D plane using stereo cameras. Hence, what we proposed is a method as discussed in section F.

Instead of taking multiple realization [6] since considering practical implementation it is not more efficient hence that we introduce novel approach to reduce noise in disparity image by dilation.

Once height is there and the plane of the object is gained using LRF our focus is to implement in a mobile platform and conclude the map reconstruction.

The dimension data obtained can be used reconstruct the objects in a virtual reality. To create a 3D model of the environment accurately. The identified dimensions can be used compute volume surface area, and they can be used for through analysis of the objects to find the density, mass and many other materialistic properties.

Today most of the areas are using depth extraction methods by stereo cameras with image processing. Proposed method is to take disparity using low cost cameras and dilation method is used to remove noise in disparity.

### A. Camera model identification

The cameras have its inherent distortions and to find the depth accurately through cameras its intrinsic parameters should be identified and distortions should be corrected. The simple web cameras that has been used can be conveniently modelled d as pinhole cameras. In pinhole camera model a scene view is formed by projecting 3D points unto the image plane using a perspective transformation as,

$$sm' = A(R|t)M'$$

Or,

$$S\begin{bmatrix} u \\ v \\ 1 \end{bmatrix} = \begin{bmatrix} f_x & 0 & c_x \\ 0 & f_y & c_y \\ 0 & 0 & 1 \end{bmatrix}\begin{bmatrix} r_{11} & r_{12} & r_{13} & t_1 \\ r_{21} & r_{22} & r_{23} & t_2 \\ r_{31} & r_{32} & r_{33} & t_3 \end{bmatrix}\begin{bmatrix} X \\ Y \\ Z \\ 1 \end{bmatrix}$$

Where (X, Y, Z) is the coordinates of a 3D point in the world coordinate space. (u, v) are the coordinates of the projection point in pixels, is a principal point that is usually at the image centre, A is a camera matrix and are the focal lengths expressed in pixel units.

Real lenses usually have distortions, without modeling them it is impossible to calculate any accurate measurement from cameras. For the pinhole camera, following distortions are possible. Radial distortions (Pincushion distortion & Barrel distortion.) Tangential distortions

To identify these errors and to rectify the errors a calibration process is needed. The above-mentioned distortions can be mathematically modeled as follows. Here, $k_1$, $k_2$, $k_3$, $k_4$, $k_5$, $k_{6......,}$ $k_{14}$ are radial distortion coefficients and $P_1$, $P_2$ are tangential distortion coefficients.

$$\begin{bmatrix} x \\ y \\ z \end{bmatrix} = R\begin{bmatrix} X \\ Y \\ Z \end{bmatrix} + t$$

$$x' = \frac{x}{z}$$

$$y' = \frac{y}{z}$$

$$x'' = x'\frac{1 + k_1r^2 + k_2r^4 + k_3r^6}{1 + k_4r^2 + k_5r^4 + k_6r^6} + 2P_1x'y' + P_2(r^2 + 2x')$$

$$y'' = y'\frac{1 + k_1r^2 + k_2r^4 + k_3r^6}{1 + k_4r^2 + k_5r^4 + k_6r^6} + 2P_2x'y' + P_1(r^2 + 2y')$$

Where,

$$r^2 = x'^2 + y'^2$$

$$u = f_x x'' + C_x$$

$$v = f_y y'' + C_y$$

### B. Camera calibration

Camera calibration was done in two parts. First individual cameras were calibrated and the error parameters and Camera parameters were found. Then both cameras were calibrated as a pair to find the stereo parameters of the setup. For the calibration process a checkerboard with known dimensions was used as shown in figure 1.

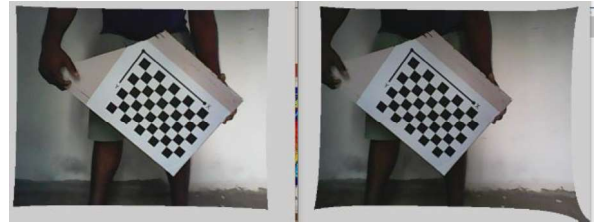

FIGURE 1: Camera calibration checker board images (left & right)

From Camera calibrations following parameters were found.

- Radial distortion: Correction matrix for radial distortion
- Tangential distortion: Correction matrix for Tangential distortion
- Focal length: Focal length of camera
- Principle point: Principal Point of the image plane

From stereo setup calibration following parameters were found

- Rotation of camera 2
  Rotation of image plane of the camera 2 with respect to camera 1
- Translation of camera 2
  Distance from the principal point of the camera 1 to principal point of the camera 2
- Fundamental Matrix
  The fundamental matrix is a relationship between any two images of the same scene that constrains where the projection of points from the scene can occur in both images.

The rotation matrix of camera two with respect to the camera one was found as,

| | | |
|---|---|---|
| 1.0000 | 0.0027 | -0.0036 |
| -0.0027 | 1.0000 | 0.0016 |
| 0.0036 | -0.0016 | 1.0000 |

And the translation of camera two with respect to camera one was found as,

-93.1032   1.2802   0.1104

Fundamental Matrix

| | | |
|---|---|---|
| -0.0000 | -0.0000 | 0.0016 |
| 0.0000 | 0.0000 | 0.1269 |
| -0.0018 | -0.1274 | 0.6250 |

Essential Matrix

| | | |
|---|---|---|
| -0.0048 | -0.4414 | 1.0334 |
| 0.1125 | 0.1493 | 93.1061 |
| -1.2801 | -93.1021 | 0.1459 |

TABLE I
CAMERA PARAMETERS

| Parameter | Camera L | Camera R |
|---|---|---|
| Radial distortion | [0.0644 -0.2494 -0.6359] | [-0.0101 0.3192 -2.2780] |
| Tangential distortion | [6.9078e-04 -0.0011] | [-0.0048 -0.0035] |
| Focal length | [729.9077 729.4782] | [733.1340 732.8580] |
| Principle point | [322.5457 226.0965] | [303.3310 224.2269] |

Found parameters were used to correct the distortions in the images were aligned horizontally through the process of rectification.

*C. Disparity map generation*

In figure 2, I represent the image plane and c denotes the principal point of image plane. (X, Y, Z) are

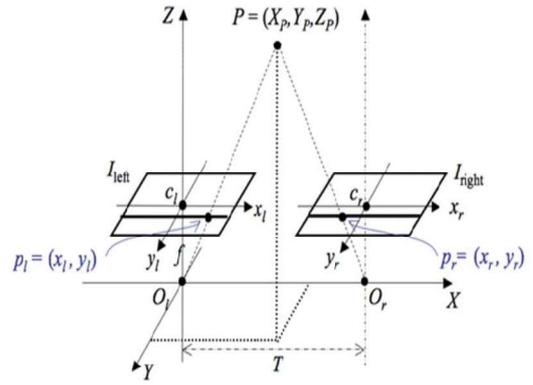

FIGURE2: Meaning of disparity

Cartesian coordinates. P denotes the real-world point and p denotes the images of that point. Subscript l and r denotes left and right. From geometry, a relationship between real world object points and image points can be derived as,

$$Z_p = f\frac{T}{d},$$
$$X_p = x_l\frac{T}{d},$$
$$Y_p = y_l\frac{T}{d},$$

Here d stands for disparity which can be mathematically represented as,

$$d = x_l - x_r,$$

Disparity in each pixel is namely disparity map which is most of the time generated by Semi Global Block Matching (SGBM) [1]. It has been used as the basis to create a disparity map by using stereo images shown in figure 3.

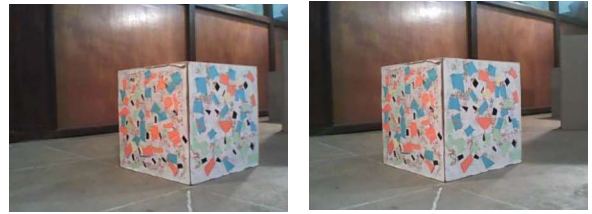

FIGURE3: Stereo images of cube

Disparity depends on the various parameters. An experiment was conducted to find those parameters as below.

For that image pairs were captured by increasing the distance from baseline up to 1m. Also, the texture pattern of the surfaces was varied.

The disparity was found and tuned manually to an optimum state. The parameter values were recorded.

Through the experiment following were concluded,

- Obviously, the no of disparities depends on the distance from the baseline, it should decrease when going further away from the baseline.
- As the LRF is available to measure the actual distance from the baseline, adjusting the parameters such that the relative depth of the object is observable corresponding to its surfaces would be sufficient.

- The Block size depends heavily on the texture of the object surface.
- Glare on the object surfaces affects the disparity much.
- The glare on the ground plane too affect it.
- Also, it is hard to compute the disparity with unicolor objects.

Hence it is Recommended that the objects and the ground plane have matt surfaces with a texture variation. Recommended distance from baseline is around 0.75m for the camera setup used in this work. But it may depend on the scale of the object.

With the results of disparity as shown in figure 4, following adjustments were made, test environment which was a cube (30 cm x 30cm x30cm) and the distance to cube was 70cm away from the camera which had a height of 8.5 similar to the mobile platform height since it was planned to mount the camera on a mobile robot. Distance from the camera to nearest edge of the object is 0.70 m

Figure 4(a) shows disparity map without texture surface with good light condition, figure 4(b) shows disparity map with texture surface with poor light condition and figure 4(c) disparity map without texture surface with good light condition while figure 4(c) shows disparity map with texture surface with good light condition. From all these observations, it is proved that disparity depends on various factors.

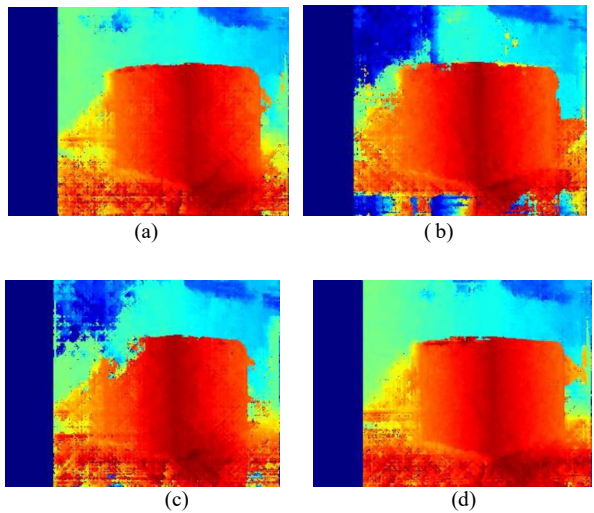

FIGURE4: Disparity maps with various test conditions

### D. Disparity map smoothing

Map of disparity was generated using SGBM and due to various noise components disparity map is not accurate. Hence, dilation was used to remove noise. Although dilation is technique used to smoothen raw camera images in simple image processing, it was applied to smoothen the disparity image due to the internal process of dilation was filling out the pixels with a large intensity deviations from the surrounding pixels. As the noise found in the disparity images were of that region, dilation was suitable. Dilation of A with a kernel of B is defined by,

$$(A \oplus B) = \bigcup_{b \in B} A_b$$

$A_b$ is the translation of $A$ by $b$.

### E. The Laser Range Finder(LRF) for 2D map generation

The LRF, Neato laser sensor driven by the Xv-11 LIDAR controller. The laser sensor projects a pulse stream of lasers around and measures the time it takes to reflect back [7]. The distance to the reflection point can be accurately calculated (to 1mm) from the time of flight. The error characteristics of the LRF is shown in the figure 5. The sensor then encodes the distance data into a packet and forward it to the LIDAR controller. The data packets are then decoded in the controller and relayed through the serial port up to a host device (a PC or a single board computer like raspberry pi) As the percentage error is small, in distances less than 3m, the LRF measurement can be used as accurate estimation of distance.

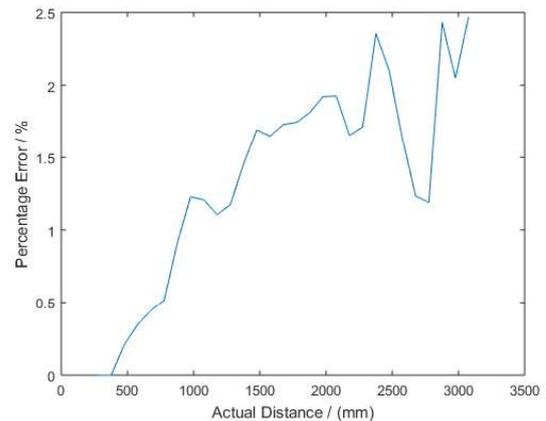

FIGURE5: The error characteristics of LRF

To acquire the distance data from the LRF a separate algorithm was developed. Because the LRF measure the distance with lesser variance in the range of 0-3m, only distances up to 3m are considered. Also, there may be errors due to reflectivity effects. But many of them could be rectified by sampling through multiple realizations. Here it was about five realizations, considering speed of operation and the power consumption. As shown in figure 6 accurate 2D map can be generated using LRF with multiple realizations.

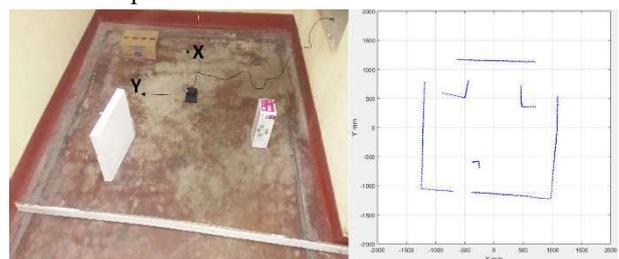

FIGURE6: 360⁰ sweep LRF data

*F.* Height extraction

From the disparity map we obtained next we focused on finding the maximum intensity from that since it is consisted with the nearest edge of the object as shown in figure 7. But the same intensity values present in the floor also affects. Hence the images were captured from a constant distance and it was found from the pixel values.

The canny edge detection method was used [8] to draw the bounding rectangular contour and obtain the height of the object in pixels as. And since top minimum row index, of the pixels with maximum intensity was known from simple calculation as below the constant was obtained in pixels. (Constant = 53 pixels). The bounding rectangle, found through canny detector, was varying for different images of the same object and the scene. Hence, another method was proposed to finds the height of the object through disparity image

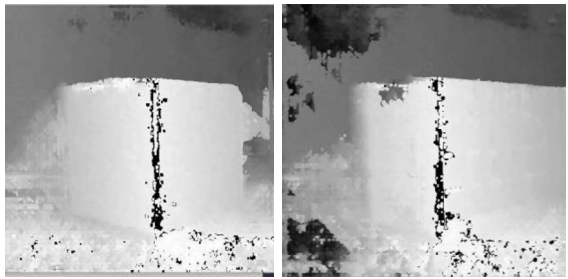

FIGURE7: Edge detection results using maximum intensity value of disparity map

### III. HEIGHT EXTRACTION METHOD USING MAXIMUM INTENSITY VALUE.

First the pixel intensity values were accessed and a threshold for filtering out the maximum intensity was identified. Hence it was used to filter out the maximum intensity value of disparity map. Once it was found, the whole disparity image was analysed row wise until first pixel with highest intensity is found. And its column index was recorded. Since the total height of the image and the number of pixels from bottom to lower end of the object are known. Once the maximum intensity points' column index is identified the height of the object can be easily calculated as shown in figure 8.

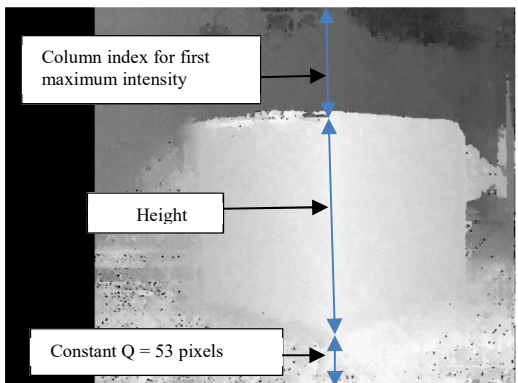

FIGURE8: Edge detection results using maximum intensity value of disparity map

### IV. PIXEL −REAL DISTANCE RELATION

Since same distance and same camera height were used all the time, they are valid for this experiment also. CANNY edge detector was used as in figure 9 to identify the width of the cube (30x30x30) from corner to corner in pixels.

Width of the cube           = 362pixels
Actual distance between corners  = 42cm
Distance represented by 1 pixel   = 0.116 cm

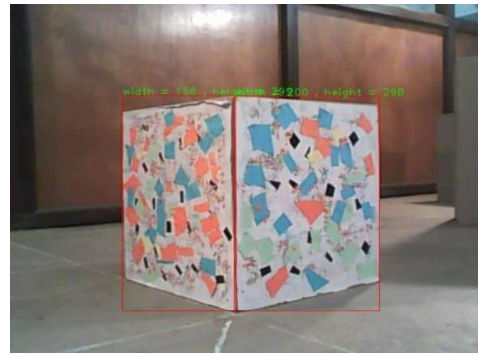

FIGURE9: Finding the corner to corner distance using CANNY edge detector rectangle

### V. CONCLUSIONS

In this paper, a method has been proposed for extracting the object dimensions for environment mapping with low cost cameras fused with laser ranging. Considering camera calibrations, it was shifted from the traditional calibrations methods and extended up to 14 levels. Furthermore, dilation method was introduced to remove the noise in the disparity map instead of using high processing power costly methods that needed multiple realizations. Considering the x-y detail extraction of object, we have introduced LRF and from that we have simulated and emulated the 2D map construction using LRF as we discussed about it in section *E*. By using the height, we gained from the method we proposed 3D map can be generated on the top of the 2D map and at the end environment mapping can be achieved.

Hence with the 3D map it can be used to identify object types, distance to object and also this can be extending to find the material of the objects too. Instead of using high cost equipment this proposed method is economically efficient and easy to implement as we did.

The ratio between dimension in pixels and real dimensions depends on the distance from the base line. Hence a parametric relationship could be identified in between them. This can be used to extend the proposed algorithm to identifying dimensions of the objects from any measurable distance from the base line.